\documentclass[10pt,twocolumn,letterpaper]{article}

\usepackage[pagenumbers]{iccv} 
\usepackage[normalem]{ulem}
\useunder{\uline}{\ul}{}
\usepackage{multirow}
\usepackage{placeins}

\definecolor{gold}{rgb}{1.0, 0.874, 0}
\definecolor{silver}{rgb}{0.77,0.77,0.77}
\definecolor{brown}{rgb}{0.95, 0.678, 0.4}

\newcommand{\gold}[1]{\colorbox{gold}{\textbf{#1}}}
\newcommand{\silver}[1]{\colorbox{silver}{\textbf{#1}}}
\newcommand{\bronze}[1]{\colorbox{brown}{\textbf{#1}}}

\definecolor{iccvblue}{rgb}{0.21,0.49,0.74}
\usepackage[pagebackref,breaklinks,colorlinks,allcolors=iccvblue]{hyperref}

\def\paperID{7914} 
\def\confName{ICCV}
\def\confYear{2025}

\title{ZeroStereo: Zero-shot Stereo Matching from Single Images}

\author{Xianqi Wang$^{1}$, ~~Hao Yang$^{1}$, ~~Gangwei Xu$^{1}$, ~~Junda Cheng$^{1}$, ~~Min Lin$^{1}$\\
~~Yong Deng$^{2}$, ~~Jinliang Zang$^{2}$, ~~Yurui Chen$^{2}$, ~~Xin Yang$^{3,1}$\footnotemark[2]\\
[2mm]
$^1$Huazhong University of Science and Technology \quad $^2$Autel Robotics \quad $^3$Optics Valley Laboratory\\
{\tt\small \{xianqiw, haoyang2002, gwxu, jundacheng, minlin, xinyang2014\}@hust.edu.cn
}}

\begin{document}

\twocolumn[{
    \maketitle
    \begin{center}
        \phantomsection
        \label{fig:head}
        \includegraphics[width=1.0\textwidth]{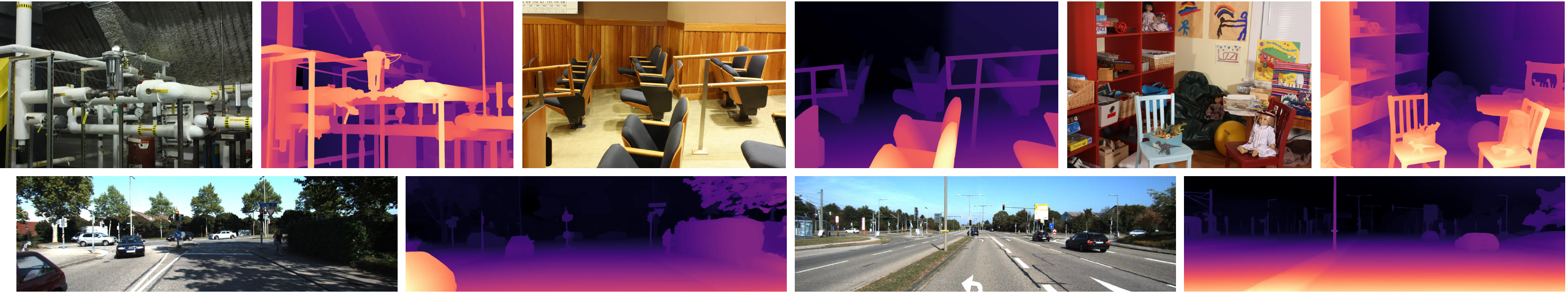}
        \captionof{figure}{Zero-shot generalization results by RAFT-Stereo~\cite{lipson2021raft} trained under our ZeroStereo pipeline.}
    \end{center}
}]



\begin{abstract}
State-of-the-art supervised stereo matching methods have achieved remarkable performance on various benchmarks. However, their generalization to real-world scenarios remains challenging due to the scarcity of annotated real-world stereo data. In this paper, we propose ZeroStereo, a novel stereo image generation pipeline for zero-shot stereo matching. Our approach synthesizes high-quality right images from arbitrary single images by leveraging pseudo disparities generated by a monocular depth estimation model. Unlike previous methods that address occluded regions by filling missing areas with neighboring pixels or random backgrounds, we fine-tune a diffusion inpainting model to recover missing details while preserving semantic structure. Additionally, we propose Training-Free Confidence Generation, which mitigates the impact of unreliable pseudo labels without additional training, and Adaptive Disparity Selection, which ensures a diverse and realistic disparity distribution while preventing excessive occlusion and foreground distortion. Experiments demonstrate that models trained with our pipeline achieve state-of-the-art zero-shot generalization across multiple datasets, with only a dataset volume comparable to Scene Flow. Code: \textcolor{magenta}{https://github.com/Windsrain/ZeroStereo}.
\end{abstract}

\vspace{-10px} 
\section{Introduction}
\label{sec:intro}

Stereo matching is a fundamental task in computer vision that estimates depth information by identifying corresponding points between stereo image pairs. By computing the disparity between matched pixels, stereo matching enables 3D scene reconstruction, which is essential for applications such as autonomous driving and robotic perception.

With the advancement of deep learning, stereo matching has shifted from traditional handcrafted feature-based approaches to data-driven methods~\cite{chang2018pyramid, lipson2021raft, xu2022attention, xu2023iterative, xu2023accurate, wang2024selective, cheng2024coatrsnet, cheng2025monster}. While deep learning-based models achieve impressive performance on standard benchmarks, they struggle to generalize to real-world scenarios due to the scarcity of annotated real-world stereo data~\cite{tosi2025survey}. Most models rely on synthetic datasets~\cite{mayer2016large, wang2020tartanair} or limited real-world datasets~\cite{scharstein2014high, schops2017multi} which fail to cover the full diversity of real-world environments. Several approaches have been proposed to mitigate this challenge.

One direction involves learning domain-invariant feature representations from synthetic data~\cite{zhang2020domain, liu2022graftnet, chuah2022itsa, chang2023domain, rao2023masked}. However, a domain gap persists due to fundamental differences between synthetic and real-world data distributions. Another approach leverages self-supervised learning~\cite{tonioni2019learning, tonioni2019real}, using photometric loss~\cite{godard2019digging} as a proxy supervision signal on unlabeled stereo images. However, this method struggles with occlusions, ghosting artifacts, and ambiguities in ill-posed regions, while large-scale collection of high-quality stereo image pairs remains non-trivial.

In recent years, view synthesis techniques~\cite{mildenhall2021nerf, pumarola2021d} have emerged as a promising approach to self-supervised stereo matching. These methods generate pseudo stereo images and corresponding disparity labels from single images or NeRF-rendered scenes. Early strategies~\cite{luo2018single, watson2020learning} employ monocular depth estimation~\cite{li2018megadepth, godard2019digging, ranftl2020towards} to derive pseudo disparity labels, followed by forward warping to synthesize the right image. However, this approach struggles with occluded regions, where missing pixels are typically filled using neighboring pixels~\cite{luo2018single} or random backgrounds~\cite{watson2020learning}, resulting in structural inconsistencies. To address this, NeRF-Stereo~\cite{tosi2023nerf} has been proposed to generate stereo images from NeRF-rendered scenes. It leverages an implicit 3D representation, enabling it to synthesize occluded regions during rendering, rather than relying on post-processing heuristics. Additionally, it introduces Ambient Occlusion~\cite{muller2022instant} as a confidence measure to enhance the reliability of pseudo disparity. However, NeRF-Stereo requires multi-view inputs for scene reconstruction, limiting its flexibility compared to single-image-based methods. Moreover, NeRF’s reconstruction quality for distant objects is often suboptimal, leading to degraded stereo generation in large-scale outdoor environments~\cite{cheng2024adaptive}.

To overcome these challenges, we propose ZeroStereo, a novel stereo image generation pipeline for zero-shot stereo matching. Inspired by Marigold~\cite{ke2024repurposing}, we hypothesize that modern diffusion models, pre-trained on large-scale image datasets, can be effectively adapted for stereo matching. However, directly applying existing diffusion inpainting models is insufficient for stereo generation, as standard inpainting tasks do not account for the complex and structured occlusion patterns in stereo pairs. To address this, we fine-tune a diffusion inpainting model specifically for stereo image synthesis, ensuring it can handle the diverse and irregular inpainting masks encountered in occluded regions. This enables our method to recover missing background details more accurately, significantly preserving semantic consistency compared to previous heuristic filling approaches. In addition to high-quality image synthesis, training stability is another key factor in stereo matching. To mitigate the impact of unreliable pseudo disparities, we introduce Training-Free Confidence Generation, which derives confidence directly from a monocular depth estimation model. Furthermore, we propose Adaptive Disparity Selection, which dynamically adjusts the disparity distribution to prevent excessive occlusions and foreground distortions. By ensuring a wider yet realistic disparity range, this component enhances the model's ability to generalize across diverse scenarios.

By integrating these components, ZeroStereo enables efficient and high-quality stereo image generation, leading to state-of-the-art zero-shot stereo matching. Remarkably, our method achieves this performance with a dataset volume comparable to Scene Flow~\cite{mayer2016large}, demonstrating its ability to generate highly effective training data without requiring large-scale real-world stereo pairs.

Our main contributions can be summarized as follows:

\begin{itemize}
\item We propose a novel stereo image generation pipeline ZeroStereo for zero-shot stereo matching, including a fine-tuned diffusion inpainting model adapting for complex inpainting masks in stereo matching.
\item We propose Training-Free Confidence Generation and Adaptive Disparity Selection to improve stereo training stability and enhance disparity diversity.
\item We demonstrate that models trained with our pipeline achieve state-of-the-art zero-shot generalization performance using only a synthesized dataset volume comparable to Scene Flow.
\end{itemize}
\section{Related Work}
\label{sec:relate}

\textbf{Deep Stereo Matching.} The advancement of deep learning has significantly improved stereo matching. Early methods~\cite{chang2018pyramid, shen2021cfnet, cheng2022region}, such as DispNet~\cite{mayer2016large} and GC-Net~\cite{kendall2017end}, employed CNNs to construct cost volumes over a predefined disparity range. More recently, iterative refinement-based methods~\cite{lipson2021raft, xu2022attention, xu2023iterative, xu2025igev++}, inspired by RAFT~\cite{teed2020raft}, have been introduced to iteratively update disparity predictions, improving accuracy and robustness. Additionally, transformer-based models~\cite{li2021revisiting, guo2022context} leverage self-attention mechanisms to capture long-range feature dependencies, enabling more effective cost volume aggregation.

\textbf{Zero-shot Generalization in Stereo Matching.} Despite these advancements, deep stereo models often struggle with generalization to real-world scenarios. DSMNet~\cite{zhang2020domain} addresses domain shifts by introducing domain normalization layers and non-local graph-based filters to enhance feature robustness. GraftNet~\cite{liu2022graftnet} improves generalization by incorporating pre-trained features from large-scale datasets, while ITSA~\cite{chuah2022itsa} mitigates shortcut learning using an information-theoretic approach. Inspired by masked representation learning, Rao et al.~\cite{rao2023masked} propose a masking-based strategy to enhance stereo feature learning. Another line of research focuses on self-supervised learning using unlabeled images. Luo et al.~\cite{luo2018single} pioneers single-view stereo training on the KITTI dataset, while MfS-Stereo~\cite{watson2020learning} generates stereo pairs from monocular images to enable training without ground-truth disparities. NeRF-Stereo~\cite{tosi2023nerf} introduces NeRF to generate stereo images from 3D scene reconstructions.

\textbf{Diffusion Models for Image Synthesis.} Denoising Diffusion Probabilistic Models (DDPMs)~\cite{ho2020denoising} have demonstrated success in image synthesis by progressively refining images through a denoising process. Latent Diffusion Models (LDMs)~\cite{rombach2022high} further improve efficiency by performing diffusion steps in a lower-dimensional latent space. ControlNet~\cite{zhang2023adding} extends these models by introducing spatial conditioning mechanisms for better control over generated content. RePaint~\cite{lugmayr2022repaint} proposes an inpainting method based on pre-trained DDPMs, showcasing the effectiveness of diffusion models in restoring missing visual details.
\begin{figure*}[t]
    \centering
    \includegraphics[width=1.0\textwidth]{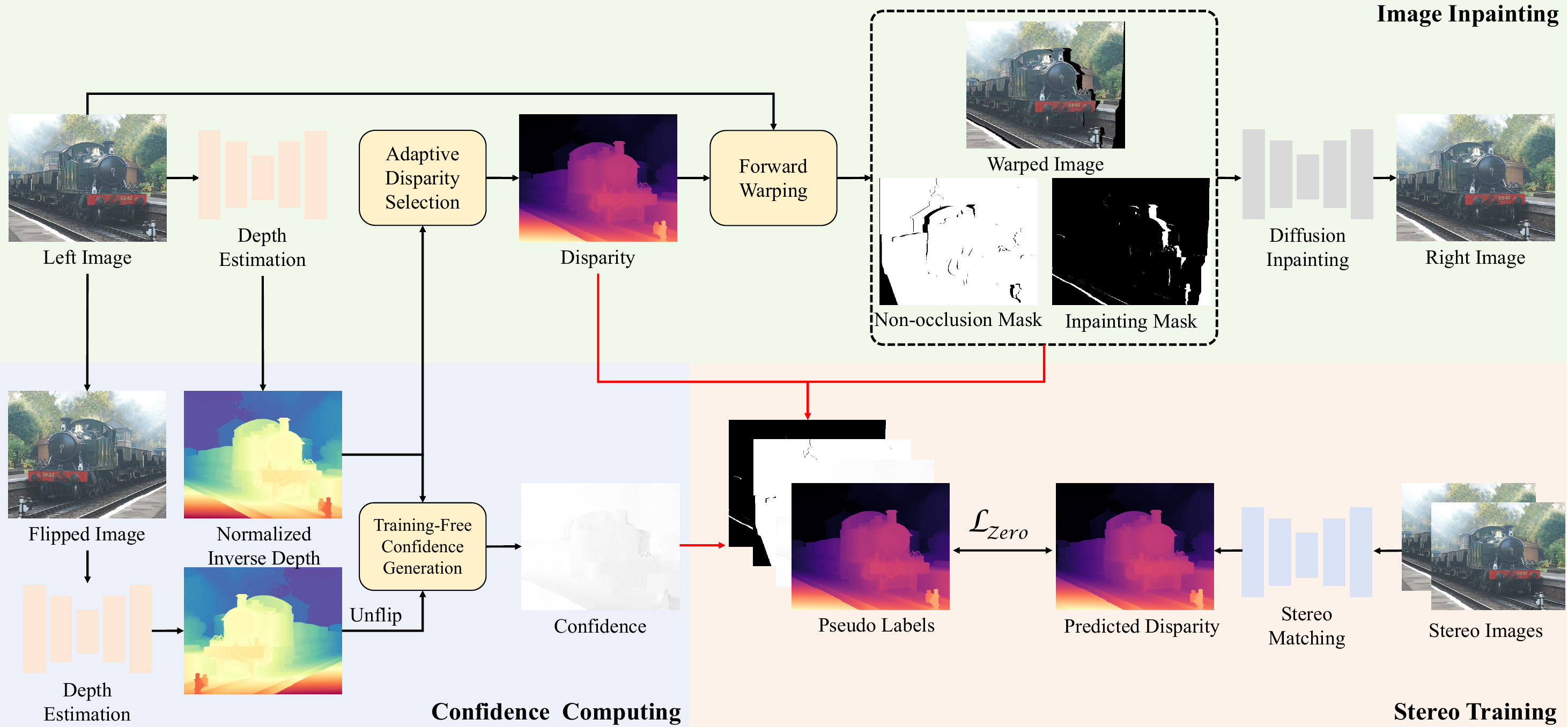}
    \caption{Overview of ZeroStereo. Given a left image, a monocular depth estimation model infers the normalized inverse depth. Our Training-Free Confidence Generation (Sec. \ref{subsec:confidence}) and Adaptive Disparity Selection (Sec. \ref{subsec:adaptive}) modules extract the confidence and pseudo disparity. Forward warping is then applied to generate a warped image and corresponding masks, which are processed by a diffusion inpainting model (Sec. \ref{subsec:inpaint}) to synthesize the right image. The final stereo images and pseudo labels are used for stereo training (Sec. \ref{subsec:stereo}).}
    \label{fig:overview}
    \vspace{-10px}
\end{figure*}

\section{Method}
\label{sec:method}

In this section, we present the overview of our ZeroStereo pipeline (Fig. \ref{fig:overview}) and details of our proposed modules.

\subsection{Overview}

Given a single image as the left image $\mathbf{I}_{l}$, we first obtain a normalized inverse depth map $\mathbf{D}$ using a monocular depth estimation model (we use Depth Anything V2~\cite{yang2024depth}, referred to as DAv2). This depth map is then converted into a pseudo disparity map $\mathbf{d}$ via our Adaptive Disparity Selection (ADS) module. Using the forward warping technique from ~\cite{watson2020learning}, we generate a warped image $\Tilde{\mathbf{I}}_{r}$, a non-occlusion mask $\mathbf{M}_{noc}$ (pixels only visible in $\mathbf{I}_{l}$), and an inpainting mask $\mathbf{M}_{inp}$ (pixels invisible in $\mathbf{I}_{l}$). $\Tilde{\mathbf{I}}_{r}$ and $\mathbf{M}_{inp}$ are then processed by a fine-tuned diffusion inpainting model to synthesize a high-quality right image $\mathbf{I}_{r}$. To improve training stability, we introduce the Training-Free Confidence Generation (TCG) module, which computes confidence $\mathbf{C}$. Finally, the synthesized stereo image pairs and associated pseudo labels are used to train stereo matching models.

\subsection{Image Inpainting}
\label{subsec:inpaint}

We fine-tune a diffusion inpainting model based on Stable Diffusion V2 Inpainting (SDv2I)~\cite{rombach2022high}. Although the pre-trained inpainting model can be directly applied, it is not specifically designed for stereo image synthesis. There exist differences between standard image inpainting and image inpainting in stereo matching.

First, there is no explicit textual guidance for inpainting. As a text-to-image model, SDv2I is trained on both text-conditioned and unconditioned data. However, no reliable textual prompt effectively directs the model to inpaint occluded regions in stereo matching. Second, unlike standard image inpainting, which typically restores or replaces specific objects or regions, occlusion masks in stereo matching exhibit diverse and irregular shapes. As a result, directly applying a pre-trained model yields suboptimal performance, necessitating fine-tuning to achieve effective results.

\textbf{Fine-tuning Protocol.} For fine-tuning, we utilize synthetic stereo datasets like Scene Flow~\cite{mayer2016large} which provide dense disparity maps as ground truth. Similar to Marigold~\cite{ke2024repurposing}, synthetic data is essential because it offers dense and complete ground truth, enabling per-pixel warping. Moreover, synthetic images are free from real-world noise, ensuring cleaner training data. Given a warped image $\Tilde{\mathbf{I}}_{r}$, an inpainting mask $\mathbf{M}_{inp}$, and a right image $\mathbf{I}_{R}$, we employ a frozen Variational Auto-Encoder (VAE)~\cite{kingma2013auto} to encode $\Tilde{\mathbf{I}}_{r}$ and $\mathbf{I}_{R}$ into the latent space. The inpainting mask $\mathbf{M}_{inp}$ is resized to match the latent space resolution. We then sample Gaussian noise $\mathbf{\epsilon}$ and add it to the latent right image.  Finally, these latent features and the resized inpainting mask are concatenated as input to the U-Net, which predicts noise $\Tilde{\mathbf{\epsilon}}$. The network is optimized using an L2 loss function:

\begin{equation}
    \mathcal{L}_{u} = || \Tilde{\mathbf{\epsilon}} - \mathbf{\epsilon} ||^{2}_{2}
\end{equation}

\begin{figure*}[t]
    \centering
    \includegraphics[width=1.0\textwidth]{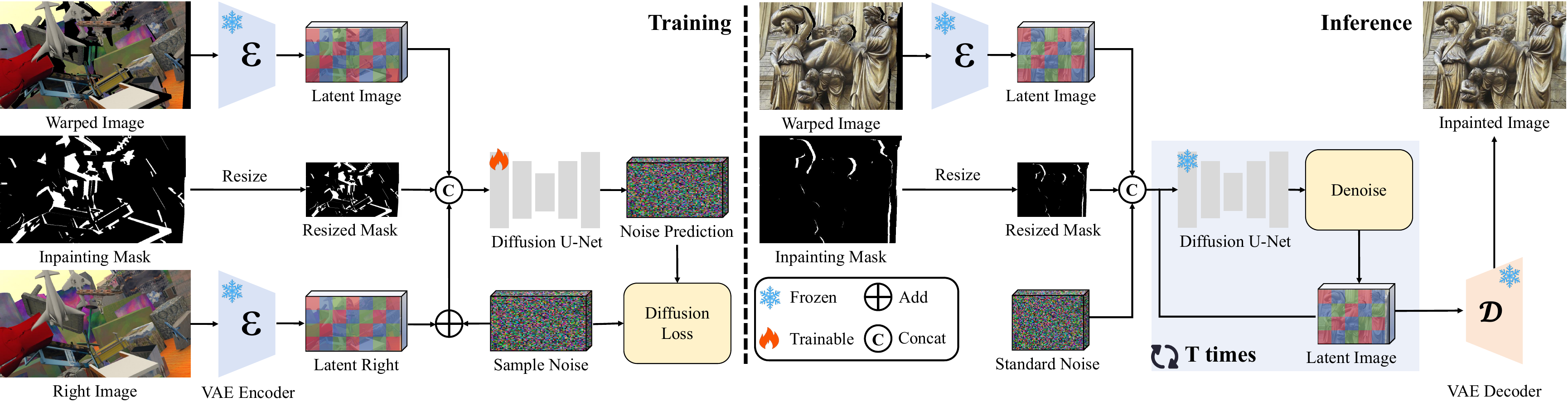}
    \caption{Overview of our diffusion inpainting protocol. During training, we freeze VAE and only fine-tune U-Net.}
    \label{fig:diffusion}
    \vspace{-10px}
\end{figure*}

\textbf{Inference Protocol.} Given a warped image $\Tilde{\mathbf{I}}_{r}$ and an inpainting mask $\mathbf{M}_{inp}$, we first encode them into the latent space. Next, we sample standard Gaussian noise to initialize the latent right image and concatenate it with the encoded inputs for the U-Net. During inference, we iteratively denoise the latent representation over $\mathbf{T}$ steps. Finally, the VAE decoder reconstructs the denoised image $\mathbf{I}_{d}$, yielding the final inpainted image $\mathbf{I}_{r}$:

\begin{equation}
    \mathbf{I}_{r} = \mathbf{M}_{inp} \odot \mathbf{I}_{d} + (1 - \mathbf{M}_{inp}) \odot \Tilde{\mathbf{I}}_{r}
\end{equation}

\subsection{Training-Free Confidence Generation}
\label{subsec:confidence}

Assessing confidence in depth predictions remains challenging for previous monocular depth estimation models, which require additional training or auxiliary modules. Some post-processing methods rely on gradients~\cite{hornauer2022gradient} or probabilistic distributions~\cite{xiang2024measuring} to estimate uncertainty.

Modern monocular depth estimation models~\cite{yang2024depth1, yang2024depth, lin2025prompting} tend to predict relative depth, often represented as inverse depth in disparity space. This representation captures the relative distances between pixels, independent of camera parameters. Therefore, when an image is flipped horizontally, the predicted relative depth between pixels is expected to remain unchanged.

Given a left image $\mathbf{I}_{l}$, we apply a horizontal flip operation $\mathbf{H(x)}$ to obtain the flipped image $\mathbf{I}_{l}'$. Both images are then processed separately by the monocular depth estimation model to generate their respective relative depth maps. Since we use DAv2~\cite{yang2024depth}, which does not impose constraints on its predicted depth range, we normalize these outputs into the normalized inverse depth maps $\mathbf{D}$ and $\mathbf{D}'$. The confidence $\mathbf{C}$ of $\mathbf{D}$ is then measured:

\begin{equation}
    \begin{aligned}
        \mathbf{u} = & 1 - | \mathbf{D} - \mathbf{H}^{-1}(\mathbf{D}') | \\
        \mathbf{C} = & \frac{\mathbf{u} - \min(\mathbf{u})}{\max(\mathbf{u}) - \min(\mathbf{u})}
    \end{aligned}
\end{equation}

\begin{figure}[t]
    \centering
    \includegraphics[width=1.0\linewidth]{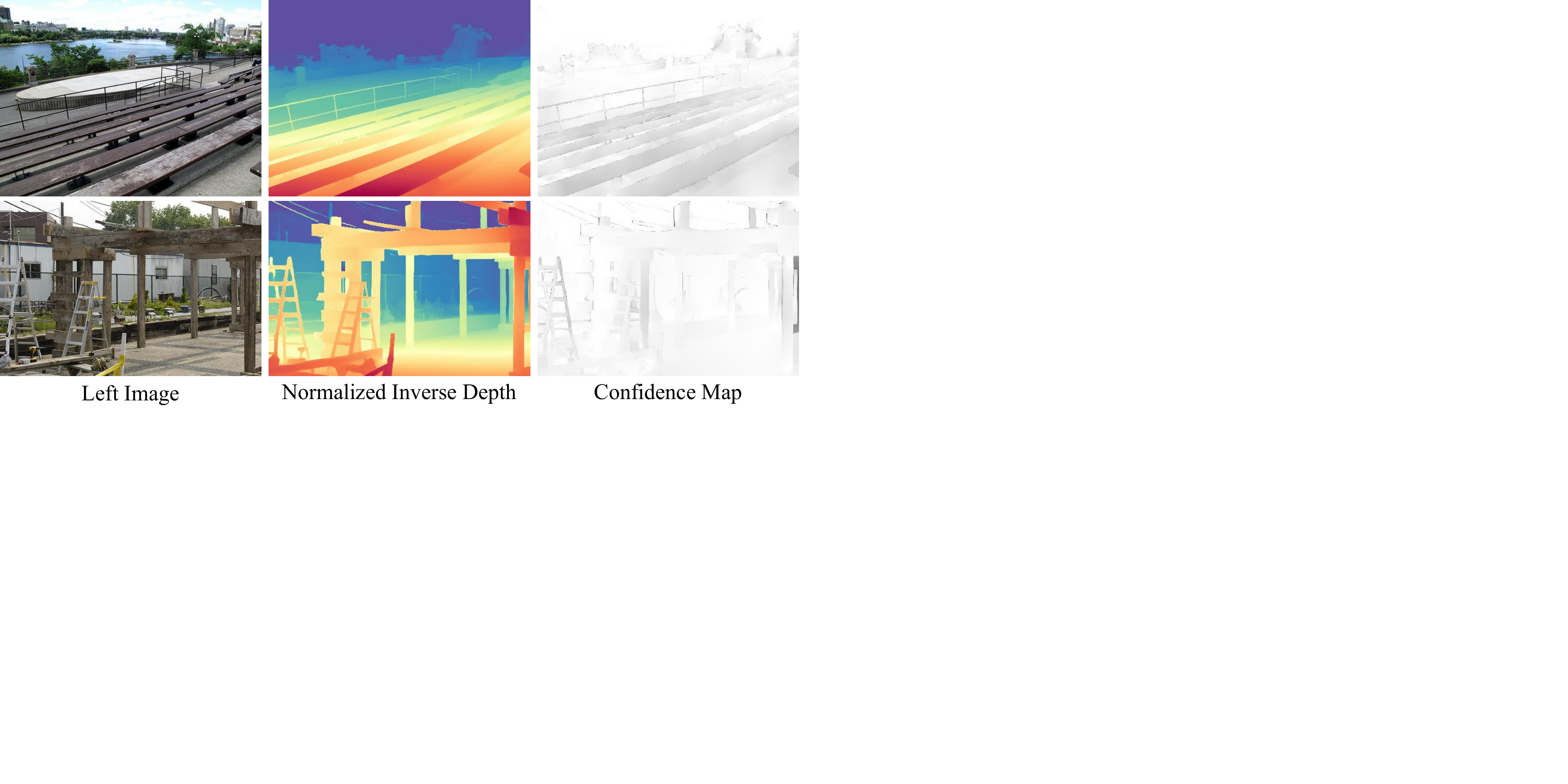}
    \caption{Visualization of confidence map.}
    \label{fig:confidence}
    \vspace{-10px}
\end{figure}

As shown in Fig. \ref{fig:confidence}, low-confidence regions typically appear along edges, textureless areas, and thin objects, which are also ambiguous in stereo matching. These unreliable labels are suppressed to mitigate their impact on learning.

\subsection{Adaptive Disparity Selection}
\label{subsec:adaptive}

The previous method, MfS-Stereo~\cite{watson2020learning}, generates disparity maps by uniformly sampling the maximum disparity value from the range $[d_{min}, d_{max}]$. However, this fixed-range approach introduces several limitations. 

First, when the image resolution is low, the disparity-to-width ratio becomes relatively large, potentially causing foreground distortion during forward warping or failure in the diffusion inpainting model due to excessive occlusion. Second, when the image resolution is high, the disparity-to-width ratio becomes relatively small, reducing the perceptible differences between the left and right images.

Therefore, we compute the disparity map $\mathbf{d}$ by multiplying $\mathbf{D}$ with the scaling factor $\mathbf{s} \cdot \mathbf{w}$, where $\mathbf{w}$ represents the image width and $\mathbf{s}$ is sampled from the distribution:

\begin{equation}
    \mathbf{s} \in \left\{
        \begin{array}{lll}
            (\mathbf{c} - 2\mathbf{r}, \mathbf{c} - \mathbf{r}), & \mathbf{p} = \mathbf{p}_{s} \\
            (\mathbf{c} - \mathbf{r}, \mathbf{c} + \mathbf{r}), & \mathbf{p} = \mathbf{p}_{c} \\
            (\mathbf{c} + \mathbf{r}, \mathbf{c} + 2\mathbf{r}), & \mathbf{p} = \mathbf{p}_{l}
        \end{array}
                 \right.
\end{equation}

where $\mathbf{c}$ is the center, $\mathbf{r}$ is the radius, and $\mathbf{p}_{i}$ ($i \in \{s, c, l\}$) is the probability. This sampling strategy ensures that most warped images maintain high-quality while occasionally introducing extreme disparity values to enhance the robustness of stereo training. Furthermore, since single-image datasets vary in resolution, this approach allows for the generation of large disparities, effectively covering a wide range of scenarios.

\begin{figure*}[t]
    \centering
    \includegraphics[width=1.0\textwidth]{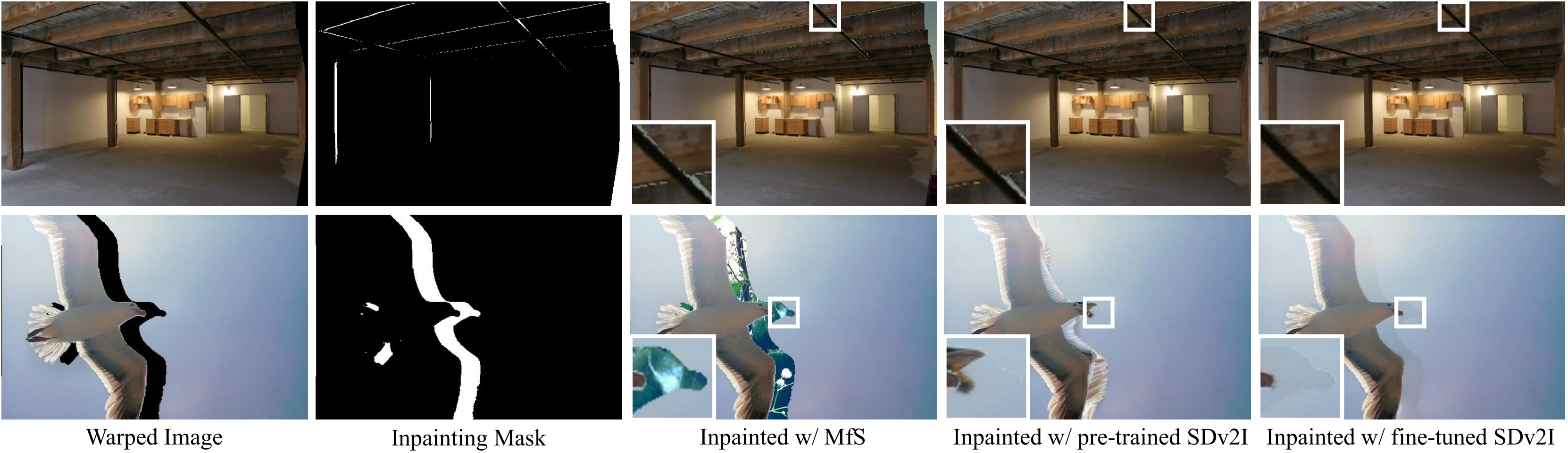}
    \caption{Visualization of different inpainting methods.}
    \label{fig:inpainting}
    \vspace{-10px}
\end{figure*}

\begin{figure*}[t]
    \centering
    \includegraphics[width=1.0\textwidth]{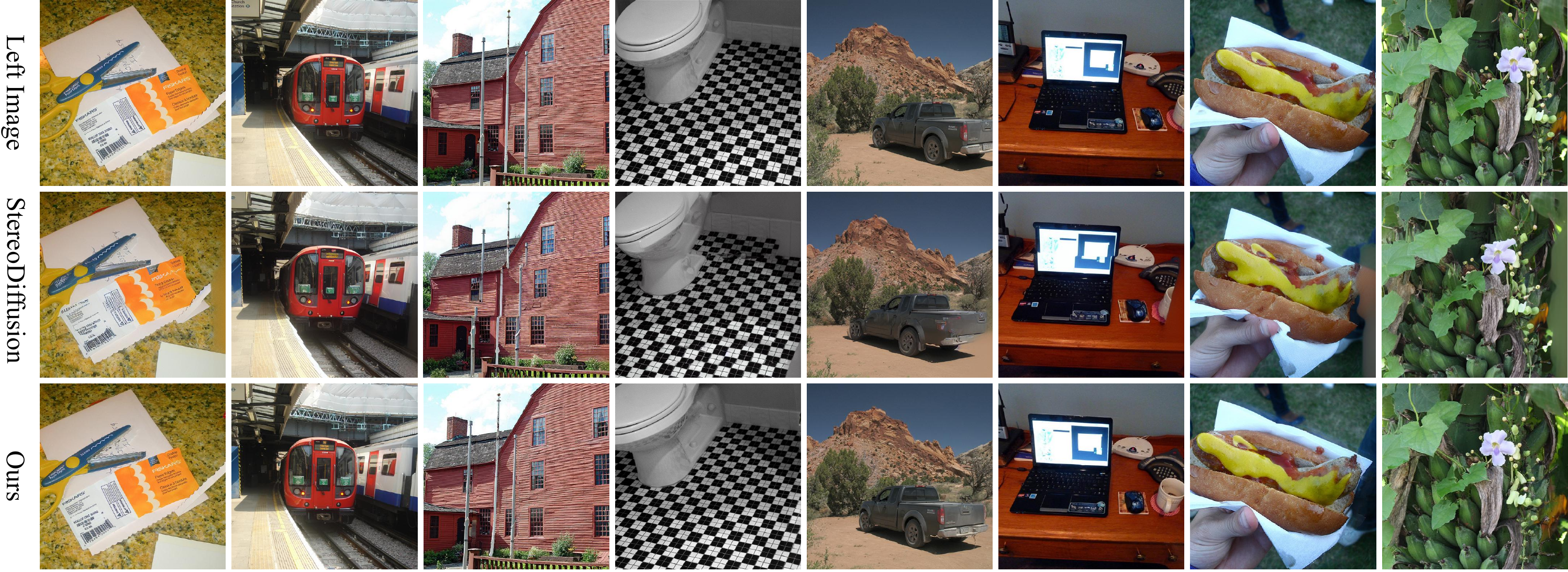}
    \caption{Visualization of StereoDiffusion~\cite{wang2024stereodiffusion} and our fine-tuned SDv2I.}
    \label{fig:stereodiffusion}
    \vspace{-10px}
\end{figure*}

\subsection{Stereo Training}
\label{subsec:stereo}

Given a stereo pair ($\mathbf{I}_{l}$, $\mathbf{I}_{r}$), a disparity map $\mathbf{d}$, a confidence map $\mathbf{C}$, a non-occlusion mask $\mathbf{M}_{noc}$, an inpainting mask $\mathbf{M}_{inp}$, and an estimated disparity map $\Tilde{\mathbf{d}}$, we train stereo matching models using our proposed ZeroStereo loss.

\textbf{Disparity Loss.} We adopt the same L1 loss as used in previous supervised stereo matching methods:

\begin{equation}
    \mathcal{L}_{d} = || \Tilde{\mathbf{d}} - \mathbf{d} ||_{1}
\end{equation}

\textbf{Non-occlusion Photometric Loss.} We backward warp $\mathbf{I}_{r}$ using the estimated disparity $\Tilde{\mathbf{d}}$ to obtain $\mathbf{I}_{l}^{r}$. The ordinary photometric loss is then computed as:

\begin{equation}
    \mathcal{L}_{p} = \beta \cdot \frac{1 - SSIM(\mathbf{I}_{l}, \mathbf{I}_{l}^{r})}{2} + (1 - \beta) \cdot || \mathbf{I}_{l} - \mathbf{I}_{l}^{r} ||_{1}
\end{equation}

$\mathbf{I}_{l}^{r}$ includes pixels inpainted by the diffusion inpainting model. To exclude these pixels, we backward warp $1 - \mathbf{M}_{inp}$ to obtain $\Tilde{\mathbf{M}}_{inp}^{r}$. Besides, $\mathbf{I_{l}^{r}}$ contains ghosting artifacts due to backward warping, which we filter using $\mathbf{M}_{noc}$. Thus, our loss is computed as follows:

\begin{equation}
    \mathcal{L}_{np} = \mathbf{M}_{noc} \odot \Tilde{\mathbf{M}}_{inp}^{r} \odot \mathcal{L}_{p}
\end{equation}

\textbf{ZeroStereo Loss.} The above two terms are summed as:

\begin{equation}
    \mathcal{L}_{Zero} = \mathbf{C} \odot \mathcal{L}_{d} + \mu \cdot (1 - \mathbf{C}) \odot \mathcal{L}_{np}
\end{equation}

\section{Experiments}
\label{sec:exp}

In this section, we present our implementation details, evaluation datasets, ablation study, and experimental results.

\begin{table*}
    \centering
    \begin{tabular}{ccccc|cc|cc|cc}
        \hline
        \multirow{2}{*}{Baseline} & \multirow{2}{*}{ADS} & \multirow{2}{*}{Inpainting} & \multirow{2}{*}{TCG} & \multirow{2}{*}{$\mathcal{L}_{Zero}$} & \multicolumn{2}{c|}{KITTI-15 All} & \multicolumn{2}{c|}{Midd-T (H) Noc} & \multicolumn{2}{c}{ETH3D Noc} \\
         &  &  &  &  & EPE & \textgreater 3px & EPE & \textgreater 2px & EPE & \textgreater 1px \\ \hline
        \checkmark &  &  &  &  & 1.52 & 4.89 & 2.71 & 8.41 & 0.25 & 2.38 \\
         & \checkmark &  &  &  & 1.24 & 4.84 & 2.28 & 7.46 & 0.28 & 2.27 \\
         &  & \checkmark &  &  & 1.44 & 4.85 & 2.34 & 7.59 & 0.23 & 1.92 \\
         &  &  & \checkmark &  & 1.09 & 4.78 & 2.13 & 7.27 & 0.27 & 2.16 \\
         & \checkmark & \checkmark &  &  & 1.06 & 4.74 & 2.26 & \textbf{6.68} & 0.23 & 2.05 \\
         & \checkmark & \checkmark & \checkmark &  & 1.05 & \textbf{4.71} & 2.18 & 7.11 & 0.23 & 2.01 \\
         & \checkmark & \checkmark & \checkmark & \checkmark & \textbf{1.04} & 4.73 & \textbf{2.09} & 7.07 & \textbf{0.22} & \textbf{1.90} \\
         \hline
    \end{tabular}
    \vspace{-5px}
    \caption{Ablation study of proposed modules trained with IGEV-Stereo~\cite{xu2023iterative}. Baseline denotes that we train the model under the pipeline in ~\cite{watson2020learning}. ADS denotes our Adaptive Disparity Selection module and TCG denotes our Training-free Confidence Generation module.}
    \label{tab:module}
    \vspace{-5px}
\end{table*}

\subsection{Implementation Details}

All experiments are implemented with PyTorch on NVIDIA RTX 4090 GPUs.

\textbf{Diffusion Inpainting Model Fine-tuning.} We utilize the Stable Diffusion V2 Inpainting (SDv2I)~\cite{rombach2022high}, disabling text conditioning and applying the DDPM noise scheduler~\cite{ho2020denoising} with 1,000 diffusion steps. We use a collection of synthetic stereo datasets, including Tartan Air~\cite{wang2020tartanair}, CREStereo Dataset~\cite{li2022practical}, Scene Flow~\cite{mayer2016large}, VKITTI 2~\cite{cabon2020virtual}, etc. Fine-tuning takes 50K steps with a batch size of 32 (gradient accumulation for 4 steps). We use the AdamW optimizer and a one-cycle learning rate schedule with a learning rate of 2e-5. Besides, we apply a crop size of $512 \times 512$ and symmetric color augmentation.

\textbf{Stereo Image Generation.} We use the DDIM scheduler~\cite{song2020denoising} and perform 50 sampling steps. Following MfS-Stereo~\cite{watson2020learning}, we sample images from Depth in the Wild~\cite{chen2016single}, COCO 2017~\cite{lin2014microsoft}, DIODE~\cite{vasiljevic2019diode}, ADE20K~\cite{zhou2017scene}, and Mapillary Vistas~\cite{neuhold2017mapillary}. We randomly sample 35K to create a dataset called MfS35K. For disparity generation, we set: $\mathbf{c} = 0.1$, $\mathbf{r} = 0.05$, $\mathbf{p}_{c} = 0.8$, and $\mathbf{p}_{s} = \mathbf{p}_{l} = 0.1$.

\textbf{Stereo Matching Model Training.} We use RAFT-Stereo~\cite{lipson2021raft} and IGEV-Stereo~\cite{xu2023iterative}. Models are trained on MfS35K with a batch size of 8 and a crop size of $384 \times 512$. We follow all the source codes' settings and train 200k steps from scratch. In addition to the data augmentation from RAFT-Stereo, we introduce Gaussian augmentation on the right image, as done in MfS-Stereo~\cite{watson2020learning}. For ZeroStereo loss, we set $\beta = 0.85, \mu = 0.1$ the same as NeRF-Stereo~\cite{tosi2023nerf}.

\begin{table} \footnotesize
    \centering
    \begin{tabular}{l|ccccc}
        \hline
        Dataset & DiW & COCO & DIODE & ADE20K & Mapillary \\ \hline
        Mean & 18.89 & 20.70 & 33.83 & 31.06 & 81.34 \\ \hline
        Max & 75.00 & 96.00 & 153.45 & 627.54 & 751.54 \\ \hline
    \end{tabular}
    \vspace{-5px}
    \caption{Disparity statistics results based on ADS.}
    \label{tab:mean}
    \vspace{-5px}
\end{table}

\begin{table} \footnotesize
    \centering
    \begin{tabular}{l|cc|cc|cc}
        \hline
        \multirow{2}{*}{Method} & \multicolumn{2}{c|}{KITTI-15 All} & \multicolumn{2}{c|}{Midd-T (H) Noc} & \multicolumn{2}{c}{ETH3D Noc} \\
         & EPE & \textgreater 3px & EPE & \textgreater 2px & EPE & \textgreater 1px \\ \hline
        Pre-trained & 1.18 & 4.77 & 2.27 & 7.21 & 0.25 & \textbf{1.90} \\
        Fine-tuned & \textbf{1.06} & \textbf{4.74} & \textbf{2.26} & \textbf{6.68} & \textbf{0.23} & 2.05 \\ \hline
    \end{tabular}
    \vspace{-5px}
    \caption{Ablation study of SDv2I.}
    \label{tab:diffusion}
    \vspace{-5px}
\end{table}

\begin{table} \footnotesize
    \centering
    \begin{tabular}{l|c|c|c}
        \hline
        Method & Resolution (px) & Memory (G) & Time (s) \\ \hline
        RePaint~\cite{lugmayr2022repaint} & $256 \times 256$ & 2.7 & 156.5 \\
        StereoDiffusion~\cite{wang2024stereodiffusion} & $512 \times 512$ & 14.6 & 31.2 \\
        Ours & $512 \times 512$ & 5.8 & 1.9 \\ \hline
    \end{tabular}
    \vspace{-5px}
    \caption{Ablation study of different synthesis methods.}
    \label{tab:time}
    \vspace{-5px}
\end{table}

\begin{table} \footnotesize
    \centering
    \begin{tabular}{l|c|c|c|c}
        \hline
        \multirow{2}{*}{Dataset} & \multirow{2}{*}{Size} & KITTI-15 & Midd-T & ETH3D \\
         &  & \textgreater 3px & \textgreater 2px & \textgreater 1px \\ \hline
        Falling Things~\cite{tremblay2018falling} & 62K & 5.14 & 5.81 & 28.63 \\
        CREStereo~\cite{li2022practical} & 200K & 6.26 & 10.91 & 2.56 \\
        Tartan Air~\cite{wang2020tartanair} & 307K & 4.91 & 5.47 & 2.69 \\
        FoundationStereo~\cite{wen2025foundationstereo} & 1106K & 4.69 & 5.11 & 2.52 \\
        MfS35K (Ours) & 35K & \textbf{4.53} & \textbf{4.45} & \textbf{2.13} \\ \hline
    \end{tabular}
    \vspace{-5px}
    \caption{Ablation study of datasets.}
    \label{tab:dataset}
    \vspace{-5px}
\end{table}

\begin{figure}[t]
    \centering
    \includegraphics[width=0.9\linewidth]{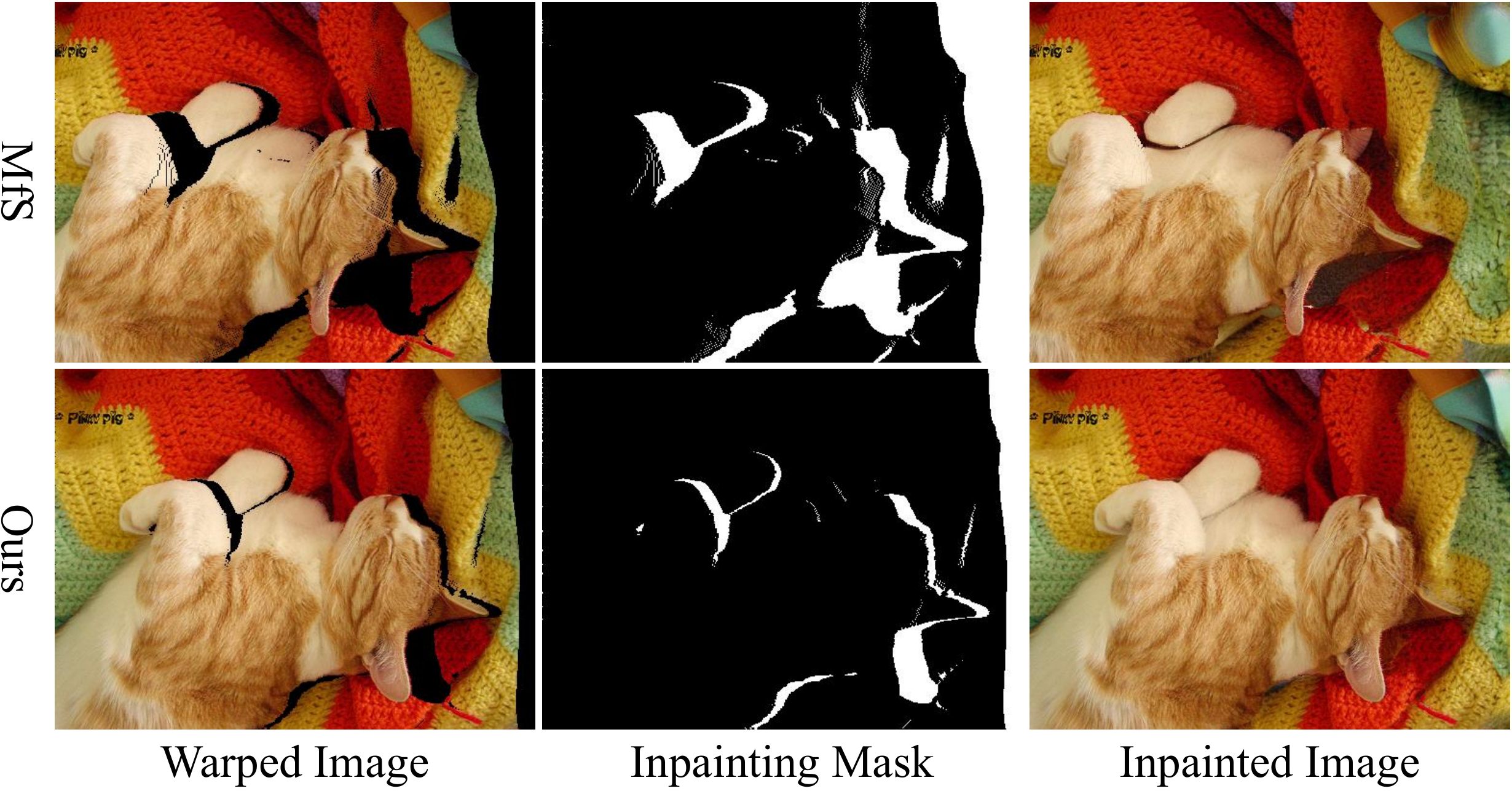}
    \caption{Visualization of different disparity selection.}
    \label{fig:range}
    \vspace{-10px}
\end{figure}

\subsection{Evaluation Datasets}

\textbf{KITTI 2015}~\cite{menze2015object} includes 200 training pairs with lidar ground truth for outdoor driving scenarios. \textbf{ETH3D}~\cite{schops2017multi} contains 27 training pairs of grayscale images, covering outdoor and indoor scenarios. For \textbf{Middlebury}~\cite{scharstein2014high}, we select the Middlebury V3 Benchmark Training Set (Midd-T), which consists of 15 training pairs for high-resolution indoor scenarios.

\begin{table*}
    \centering
    \begin{tabular}{l|cc|cccc|cc}
        \hline
        \multirow{3}{*}{Method} & \multicolumn{2}{c|}{KITTI-15} & \multicolumn{4}{c|}{Midd-T} & \multicolumn{2}{c}{ETH3D} \\
         & \multicolumn{2}{c|}{\textgreater 3px} & \multicolumn{2}{c}{F (\textgreater 2px Noc)} & \multicolumn{2}{c|}{H (\textgreater 2px Noc)} & \multicolumn{2}{c}{\textgreater 1px} \\ \cline{2-9} 
         & All & Noc & $\mathbf{D}$ \textless 192 & $\mathbf{D}$ \textless All & $\mathbf{D}$ \textless 192 & $\mathbf{D}$ \textless All & All & Noc \\ \hline
        NS-IGEV-Stereo & 5.88 & 5.58 & 15.89 & 27.91 & 8.55 & 10.67 & 4.02 & 3.58 \\
        Zero-IGEV-Stereo (Ours) & 4.73 & 4.57 & 13.03 & 26.78 & 4.73 & 7.07 & 2.64 & 1.90 \\ \hline
        NS-RAFT-Stereo\footnotemark[3]~\cite{tosi2023nerf} & 5.41 & 5.23 & 16.61 & 12.04 & 6.40 & 6.45 & 2.95 & 2.55 \\
        NS-RAFT-Stereo & 5.65 & 5.44 & 15.05 & 13.41 & 9.09 & 9.44 & 3.30 & 2.79 \\
        Zero-RAFT-Stereo (Ours) & \textbf{4.53} & \textbf{4.33} & \textbf{9.51} & \textbf{8.36} & \textbf{4.21} & \textbf{4.45} & \textbf{2.75} & \textbf{2.13} \\ \hline
    \end{tabular}
    \vspace{-5px}
    \caption{Comparison with NeRF-Stereo~\cite{tosi2023nerf}. Models are trained with the same augmentation. Exception: \textdaggerdbl{} official weights.}
    \label{tab:cmp}
    \vspace{-5px}
\end{table*}

\begin{figure*}[t]
    \centering
    \includegraphics[width=1.0\textwidth]{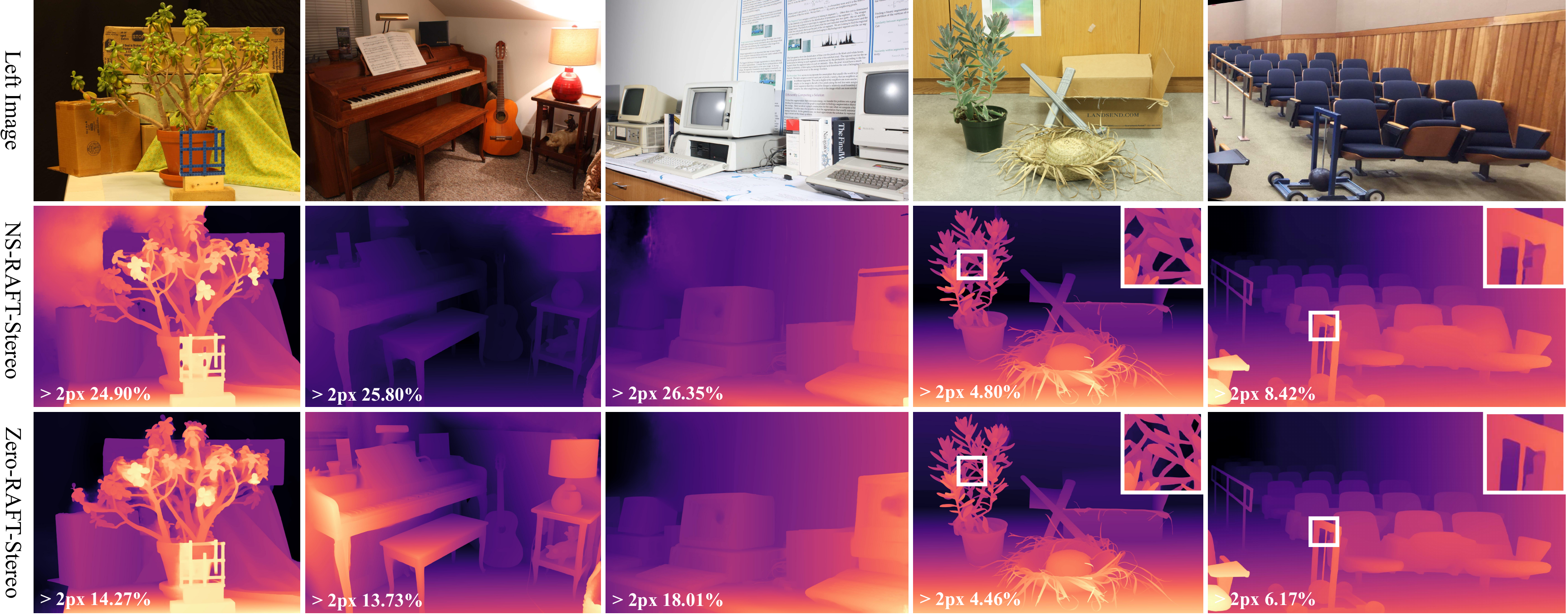}
    \caption{Visualization of Middlebury, including Midd-T, 2014 and 2021.}
    \label{fig:middlebury}
    \vspace{-10px}
\end{figure*}

\begin{table} \footnotesize
    \centering
    \begin{tabular}{l|cc|cc|cc}
        \hline
        \multirow{2}{*}{Dataset} & \multicolumn{2}{c|}{KITTI-15 All} & \multicolumn{2}{c|}{Midd-T Noc} & \multicolumn{2}{c}{ETH3D Noc} \\
         & EPE & \textgreater 3px & EPE & \textgreater 2px & EPE & \textgreater 1px \\ \hline
        Scene Flow~\cite{mayer2016large} & 1.44 & 8.17 & 2.49 & 15.81 & 0.29 & 3.79 \\ 
        NS65K~\cite{tosi2023nerf} & 1.39 & 7.60 & \textbf{2.24} & 11.99 & 0.32 & 4.58 \\ 
        MfS35K (Ours) & \textbf{1.37} & \textbf{7.41} & 2.35 & \textbf{11.92} & \textbf{0.24} & \textbf{3.38} \\ \hline
    \end{tabular}
    \vspace{-5px}
    \caption{Analysis of errors on edge (RAFT-Stereo).}
    \label{tab:edge}
    \vspace{-5px}
\end{table}

\subsection{Ablation Study}

In this section, we evaluate models with different settings to verify the effectiveness of our proposed pipeline.

\textbf{Effectiveness of proposed modules.} Tab. ~\ref{tab:module} shows the ablation study of our proposed modules. By adding ADS or TCG, we observe a notable reduction in EPE for both KITTI-15 and Midd-T. ADS improves the model's ability to handle large disparities. As shown in Tab. ~\ref{tab:mean}, the disparity range adjusts adaptively according to the dataset resolution. TCG suppresses unreliable labels, particularly in edges and textureless regions. However, adding Inpainting alone results in only a slight improvement. As shown in Fig. ~\ref{fig:range}, the large disparity ratio causes separation and distortion in the foreground, which hinders the effectiveness of the diffusion inpainting model. When ADS is combined with Inpainting, we observe a significant performance improvement.

When ADS, Inpainting, and TCG are all added, the performance consistently improves. Moreover, by introducing the final ZeroStereo Loss, the model can learn with $\mathcal{L}_{Zero}$ in low-confidence regions. This loss helps maintain the model's robustness, enabling it to achieve optimal performance across various datasets. A more detailed analysis can be found in our supplementary materials.

\textbf{Effectiveness of fine-tuned SDv2I.} As shown in Fig. ~\ref{fig:inpainting}, pixels inpainted using our fine-tuned SDv2I preserve minimal noise and maintain the semantic structure closest to the background. As shown in Tab. ~\ref{tab:diffusion}, the fine-tuned SDv2I outperforms the pre-trained model. It suggests that fine-tuning enhances the inpainting model's ability to capture the semantic structure of the background more accurately.

\textbf{Comparison with other synthesis methods.} StereoDiffusion~\cite{wang2024stereodiffusion} introduces a training-free method for generating stereo images using the pre-trained SDv2. However, the inherent inconsistency arises because the warping operation is performed in the latent space. As shown in Fig. ~\ref{fig:stereodiffusion}, StereoDiffusion suffers from structural distortions, texture inconsistencies, and poor occlusion handling, leading to unrealistic right-image generation. As shown in Tab. ~\ref{tab:time}, our fine-tuned SDv2I is significantly faster and more memory-efficient than StereoDiffusion~\cite{wang2024stereodiffusion}, while still achieving high-resolution synthesis. A more detailed discussion is available in our supplementary materials.

\textbf{Comparison with other synthetic datasets.} In recent years, many synthetic datasets with better diversity and rendering realism have been proposed. As shown in Tab. ~\ref{tab:dataset}, our MfS35K outperforms others, despite being significantly smaller. It reveals that rather than the absolute size of datasets, the diversity of scenarios is more beneficial for zero-shot generalization.

\begin{table*}
    \centering
    \begin{tabular}{lcccccccccc}
        \hline
        \multicolumn{1}{l|}{\multirow{3}{*}{Method}} & \multicolumn{2}{c|}{KITTI-15} & \multicolumn{6}{c|}{Midd-T} & \multicolumn{2}{c}{ETH3D} \\
        \multicolumn{1}{l|}{} & \multicolumn{2}{c|}{\textgreater 3px} & \multicolumn{2}{c}{F (\textgreater 2px)} & \multicolumn{2}{c}{H (\textgreater 2px)} & \multicolumn{2}{c|}{Q (\textgreater 2px)} & \multicolumn{2}{c}{\textgreater 1px} \\ \cline{2-11} 
        \multicolumn{1}{l|}{} & All & \multicolumn{1}{c|}{Noc} & All & Noc & All & Noc & All & \multicolumn{1}{c|}{Noc} & All & Noc \\ \hline
        Training Set & \multicolumn{10}{c}{Scene Flow with GT} \\ \hline
        \multicolumn{1}{l|}{DSMNet~\cite{zhang2020domain}} & 5.50 & \multicolumn{1}{c|}{5.19} & 41.96 & 38.54 & 18.74 & 14.49 & 13.75 & \multicolumn{1}{c|}{9.44} & 4.03 & 3.62 \\
        \multicolumn{1}{l|}{CFNet~\cite{shen2021cfnet}} & 5.99 & \multicolumn{1}{c|}{5.79} & 35.21 & 30.05 & 21.99 & 17.69 & 14.21 & \multicolumn{1}{c|}{10.51} & 6.08 & 5.48 \\
        \multicolumn{1}{l|}{Graft-PSMNet~\cite{liu2022graftnet}} & 5.34 & \multicolumn{1}{c|}{5.00} & 39.92 & 36.30 & 17.65 & 13.36 & 13.92 & \multicolumn{1}{c|}{9.23} & 11.43 & 10.70 \\
        \multicolumn{1}{l|}{ITSA-CFNet~\cite{chuah2022itsa}} & \silver{4.73} & \multicolumn{1}{c|}{4.67} & 34.01 & 30.14 & 16.48 & 12.32 & 12.28 & \multicolumn{1}{c|}{8.54} & 5.43 & 5.17 \\
        \multicolumn{1}{l|}{HVT-PSMNet~\cite{chang2023domain}} & \bronze{4.84} & \multicolumn{1}{c|}{\bronze{4.63}} & 40.74 & 37.60 & 15.66 & 12.55 & 10.12 & \multicolumn{1}{c|}{7.00} & 6.07 & 5.65 \\
        \multicolumn{1}{l|}{RAFT-Stereo~\cite{lipson2021raft}} & 5.47 & \multicolumn{1}{c|}{5.27} & \silver{15.63} & \silver{11.94} & 11.20 & 8.66 & 10.25 & \multicolumn{1}{c|}{7.44} & \gold{2.60} & 2.29 \\
        \multicolumn{1}{l|}{IGEV-Stereo~\cite{xu2023iterative}} & 6.03 & \multicolumn{1}{c|}{5.76} & 30.94 & 28.98 & 11.90 & 9.45 & 8.88 & \multicolumn{1}{c|}{6.20} & 4.04 & 3.60 \\
        \multicolumn{1}{l|}{NMRF-Stereo~\cite{guan2024neural}} & 5.31 & \multicolumn{1}{c|}{5.14} & 37.63 & 35.25 & 13.36 & 10.90 & 7.87 & \multicolumn{1}{c|}{5.07} & 3.80 & 3.48 \\
        \multicolumn{1}{l|}{Mocha-Stereo~\cite{chen2024mocha}} & 5.97 & \multicolumn{1}{c|}{5.73} & 30.23 & 28.26 & 10.18 & 9.45 & 7.96 & \multicolumn{1}{c|}{4.87} & 4.02 & 3.47 \\
        \multicolumn{1}{l|}{DKT-RAFT~\cite{zhang2024robust}} & 4.95 & \multicolumn{1}{c|}{4.74} & \bronze{16.05} & \bronze{9.26} & 10.18 & \bronze{6.10} & 10.39 & \multicolumn{1}{c|}{6.64} & 2.77 & 2.53 \\
        \multicolumn{1}{l|}{Former-RAFT~\cite{zhang2024learning}} & 5.18 & \multicolumn{1}{c|}{4.93} & - & - & 13.27 & 10.29 & 8.51 & \multicolumn{1}{c|}{5.61} & 3.96 & 3.50 \\ \hline
        Training Set & \multicolumn{10}{c}{Real-world data without GT} \\ \hline
        \multicolumn{1}{l|}{MfS-PSMNet~\cite{watson2020learning}} & 5.18 & \multicolumn{1}{c|}{4.91} & 26.42 & 20.91 & 17.56 & 13.45 & 12.07 & \multicolumn{1}{c|}{9.09} & 8.17 & 7.44 \\
        \multicolumn{1}{l|}{NS-RAFT-Stereo~\cite{tosi2023nerf}} & 5.41 & \multicolumn{1}{c|}{5.23} & 16.38 & 12.04 & \bronze{9.70} & 6.45 & 8.09 & \multicolumn{1}{c|}{4.85} & 2.95 & 2.55 \\
        \multicolumn{1}{l|}{Zero-IGEV-Stereo (Ours)} & \silver{4.73} & \multicolumn{1}{c|}{\silver{4.57}} & 29.47 & 26.78 & 9.71 & 7.07 & \silver{7.07} & \multicolumn{1}{c|}{\silver{4.46}} & \silver{2.64} & \gold{1.90} \\
        \multicolumn{1}{l|}{Zero-IGEV-Stereo$^{*}$ (Ours)} & 4.89 & \multicolumn{1}{c|}{4.73} & 18.83 & 14.87 & \silver{8.45} & \silver{5.54} & \gold{6.99} & \multicolumn{1}{c|}{\gold{4.38}} & 2.85 & \silver{2.00} \\
        \multicolumn{1}{l|}{Zero-RAFT-Stereo (Ours)} & \gold{4.53} & \multicolumn{1}{c|}{\gold{4.33}} & \gold{12.40} & \gold{8.36} & \gold{7.86} & \gold{4.45} & \bronze{7.24} & \multicolumn{1}{c|}{\bronze{4.50}} & \bronze{2.75} & \bronze{2.13} \\ \hline
    \end{tabular}
    \vspace{-5px}
    \caption{Zero-shot generalization benchmark. DKT-RAFT~\cite{zhang2024robust} is trained on SceneFlow~\cite{mayer2016large} and fine-tuned on Booster~\cite{ramirez2022open}. Zero-IGEV-Stereo$^{*}$ denotes that we expand the $\mathcal{L}_{d}$ supervision same as RAFT-Stereo~\cite{lipson2021raft}. We highlight \gold{first}, \silver{second}, \bronze{third} bests.}
    \label{tab:benchmark}
    \vspace{-5px}
\end{table*}

\subsection{Comparison with NeRF-Stereo}

We compare our method with NeRF-Stereo~\cite{tosi2023nerf}, a leading method for generating stereo images. As shown in Tab. ~\ref{tab:cmp}, all models are trained with the same data augmentation, except for NS-RAFT-Stereo\footnotemark[3], which uses official weights. Since the official IGEV-Stereo limits $\mathcal{L}_{d}$ supervision to a disparity of 192, we split the Midd-T table by disparity to provide a more detailed evaluation.

Re-training NS-RAFT-Stereo with our data augmentation shows no improvement, confirming that the gains are not solely due to augmentation. Zero-RAFT-Stereo outperforms NS-RAFT-Stereo\footnotemark[3] by over 20\%, with only a 10\% drop in Midd-T (F) for $\mathbf{D} < 192$, whereas NS-RAFT-Stereo\footnotemark[3] declines more, likely due to its dataset's imperfect reconstruction of distant objects. Fig.~\ref{fig:middlebury} highlights NS-RAFT-Stereo's failures in textureless regions, while our Zero-RAFT-Stereo shows over 40\% improvement in handling such cases. Moreover, as shown in Tab. ~\ref{tab:edge}, the model trained on MfS35K surpasses both the synthetic SceneFlow~\cite{mayer2016large} and the NeRF-based NS65K~\cite{tosi2023nerf}, achieving the lowest edge error and superior edge accuracy.

\subsection{Zero-shot Generalization Benchmark}

Following NeRF-Stereo~\cite{tosi2023nerf}, we construct a zero-shot generalization benchmark. All methods are evaluated across the entire disparity range. For Zero-IGEV-Stereo, we train two versions: one using the original code settings for disparity supervision, and the other expanding the supervised range, consistent with RAFT-Stereo~\cite{lipson2021raft}.

As shown in Tab. ~\ref{tab:benchmark}, our models demonstrate state-of-the-art zero-shot generalization performance across multiple datasets, both under the SceneFlow with ground truth (GT) and Real-world data without GT. Notably, Zero-RAFT-Stereo achieves the best or near-best results, particularly excelling in handling complex, real-world scenes. 

Zero-IGEV-Stereo$^{*}$, with an expanded supervised range of $\mathcal{L}_{d}$, shows improved results on Middlebury's large-disparity scenarios, although this leads to a slight performance trade-off on other datasets.
\section{Conclusion}
\label{sec:con}

We propose ZeroStereo, a novel stereo data generation pipeline for zero-shot stereo matching. The fine-tuned SDv2I adapts to complex inpainting masks and recovers background details. To handle unreliable pseudo labels, the TCG module leverages the spatial invariance of relative depth to compute confidence, helping to suppress uncertain labels. Besides, the ADS module generates a broader disparity distribution while avoiding foreground distortion. Finally, experiments demonstrate that our models achieve state-of-the-art zero-shot generalization performance.

\textbf{Limitations.} The fine-tuned SDv2I still struggles in some complex scenarios, and there may be occasional color inconsistencies due to fine-tuning on synthetic datasets. Furthermore, forward warping performs poorly in ill-posed regions, such as transparent areas or net-like objects.

\textbf{Acknowledgement.} This research is supported by the National Key R\&D Program of China (2024YFE0217700), National Natural Science Foundation of China (62472184, 623B2036), the Fundamental Research Funds for the Central Universities, and the Innovation Project of Optics Valley Laboratory (Grant No. OVL2025YZ005).

{
    \small
    \bibliographystyle{ieeenat_fullname}
    \bibliography{main}
}

\clearpage
\setcounter{page}{1}
\maketitlesupplementary

\begin{table} \footnotesize
    \centering
    \begin{tabular}{l|cc|cc|cc}
        \hline
        \multirow{2}{*}{Method} & \multicolumn{2}{c|}{KITTI-15} & \multicolumn{2}{c|}{Midd-T (H)} & \multicolumn{2}{c}{ETH3D} \\
                                & EPE       & \textgreater 3px      & EPE        & \textgreater 2px       & EPE     & \textgreater 1px    \\ \hline
        $\mathcal{L}_{p}$                   & 1.03          & 4.77                      & 0.90           & 4.95                       & 0.25        & 2.09                    \\
        $(1 - \mathbf{C}) \odot \mathcal{L}_{p}$         & 1.03      & 4.76                  & 0.85       & 4.81                   & 0.25    & \textbf{2.08}                \\
        $(1 - \mathbf{C}) \odot \mathcal{L}_{np}$     & \textbf{1.02}      & \textbf{4.53}                  & \textbf{0.79}       & \textbf{4.45}                   & \textbf{0.23}    & 2.13                \\ \hline
    \end{tabular}
    \vspace{-5px}
    \caption{Analysis of ZeroStereo loss (trained with RAFT-Stereo~\cite{lipson2021raft}). We discuss the loss combinations based on $\mathcal{L}_{d}$.}
    \label{tab:loss}
    \vspace{-5px}
\end{table}

\section{Details of Image Synthesis}
\label{sec:synthesis}

\textbf{Image Resolution.} The input resolution of Depth Anything V2~\cite{yang2024depth} and Stable Diffusion V2 Inpainting~\cite{rombach2022high} is constrained, which may lead to object deformation when resizing images. To address this, we apply padding operations to adjust image dimensions while preserving their original aspect ratio. For example, when using Depth Anything V2, we pad images to ensure their height and width are divisible by 14. Additionally, high-resolution images, particularly those from the Mapillary Vistas~\cite{neuhold2017mapillary}, may exceed available GPU memory during inference. To mitigate this issue, we first downscale images proportionally to half or quarter of their original resolution, perform inference, and then upscale the outputs to restore the original dimensions.

\textbf{Forward Warping.} We utilize the source code of MfS-Stereo~\cite{watson2020learning} to implement forward warping, including non-occlusion computation and depth sharpening. However, when applying a diffusion model for inpainting, we identify several challenges. First, despite advancements in monocular depth estimation, depth edges do not always align precisely with object boundaries. As a result, after forward warping, the edges of foreground objects may remain in their original positions. Second, the proximity between the inpainting mask and the warped foreground objects can mislead the diffusion model during inference. To mitigate these issues, we employ a simple yet effective approach: using the dilate function in OpenCV to inflate the pseudo-disparity map. This operation ensures that foreground objects and nearby background pixels move together during forward warping. Consequently, during inpainting, background pixels act as a buffer between the mask and the foreground, reducing misleading information. However, despite this refinement, the pre-trained diffusion model still produces ghosting artifacts and noise in many cases (Fig. ~\ref{fig:inpainting}). These artifacts can only be effectively addressed by fine-tuning the diffusion model.

\section{Loss Analysis}
\label{sec:analysis}

In Sec. ~\ref{subsec:stereo},  we introduce the non-occlusion photometric loss $\mathcal{L}_{np}$ and the weighted final loss $\mathcal{L}_{Zero}$. However, their specific impact on stereo training has not been explicitly analyzed. As shown in Tab. ~\ref{tab:loss}, the methods listed from top to bottom correspond to: (1) applying the ordinary photometric loss $\mathcal{L}_{p}$, (2) using $\mathcal{L}_{p}$ with the weight $1 - \mathbf{C}$, and (3) employing $\mathcal{L}_{np}$ with the weight $1 - \mathbf{C}$. 

\begin{table}
    \centering
    \begin{tabular}{l|c|c|c}
        \hline
        Method & AbsErr $\downarrow$ & SSIM $\uparrow$ & $\mathcal{L}_{p} \downarrow$ \\ \hline
        StereoDiffusion~\cite{wang2024stereodiffusion} & 0.082 & 0.269 & 0.323 \\
        Ours & \textbf{0.025} & \textbf{0.850} & \textbf{0.068} \\ \hline
    \end{tabular}
    \vspace{-5px}
    \caption{Reconstruction loss. We warp the synthesized right image with the pseudo disparity and compare it with the left image.}
    \label{tab:reconstruction}
    \vspace{-5px}
\end{table}

Among these, $\mathcal{L}_{p}$ alone yields the worst performance across all datasets due to the absence of balanced weighting and its inability to handle ghost artifacts and inpainting pixels. Introducing the weight $1 - \mathbf{C}$ mitigates these issues, leading to improved performance. The best results are achieved when masks are further applied to filter out ghost artifacts and inpainting pixels, highlighting the effectiveness of our proposed approach.

\section{Discussion on Synthesis Methods}
\label{sec:other}

In this section, we discuss two synthesis methods: StereoDiffusion~\cite{wang2024stereodiffusion} and AdaMPI~\cite{han2022single}.

\begin{figure*}[t]
    \centering
    \includegraphics[width=1.0\textwidth]{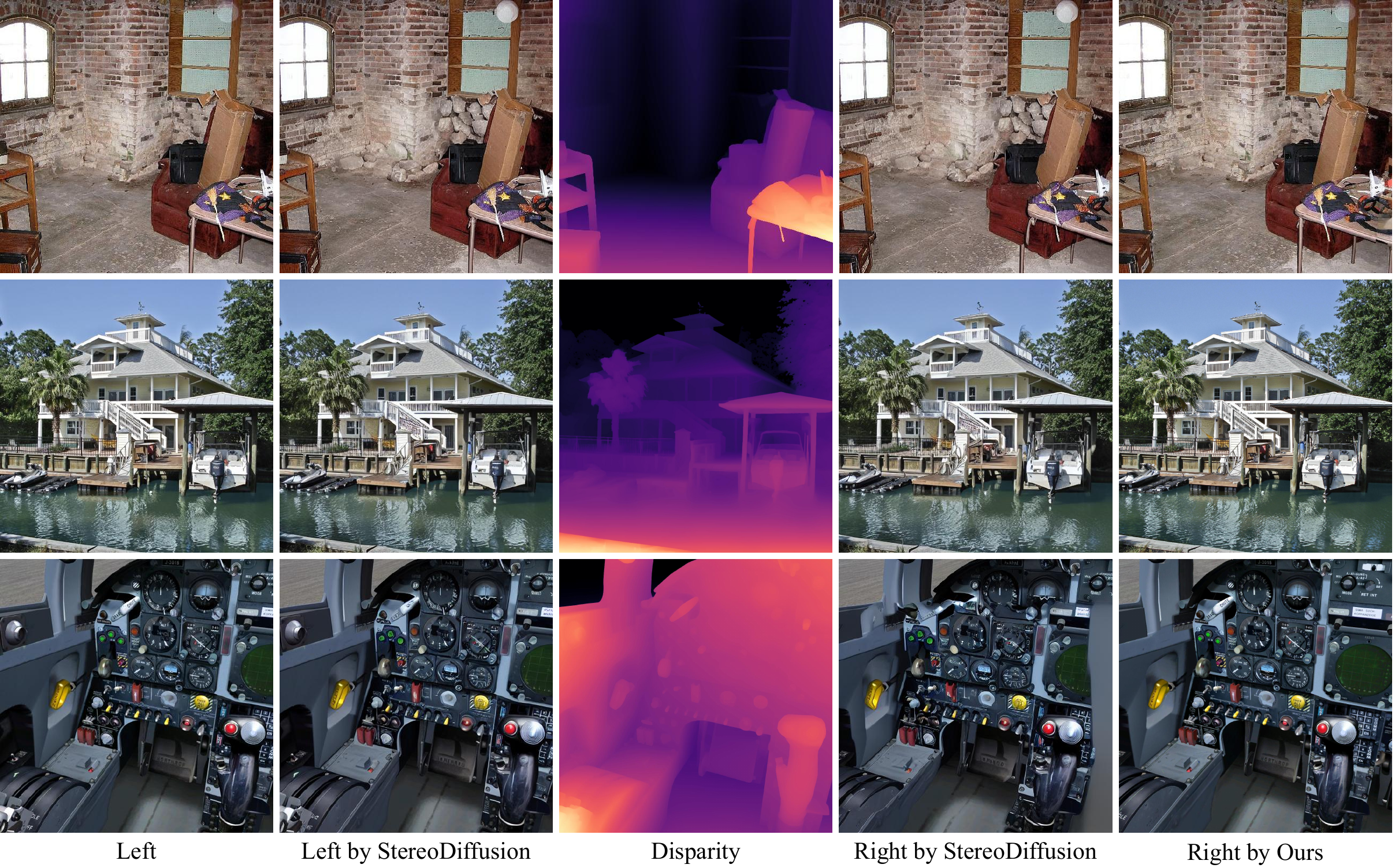}
    \caption{Visualization of StereoDiffusion~\cite{wang2024stereodiffusion}.}
    \label{fig:inversion}
    \vspace{-10px}
\end{figure*}

StereoDiffusion~\cite{wang2024stereodiffusion} is a training-free method that utilizes a pre-trained latent diffusion model to generate stereo pairs from a single image. It applies null-text inversion~\cite{mokady2023null} for image editing, first reversing the diffusion process to obtain a latent representation of the input image and then applying forward diffusion to synthesize the right view. However, this approach has notable limitations. First, inference is computationally expensive. As shown in Tab. ~\ref{tab:time}, synthesizing a $512 \times 512$ image  takes approximately 30 seconds. Second, the null-text inversion process can unintentionally modify the left image, introducing content inconsistencies. As illustrated in Fig. ~\ref{fig:inversion}, the original image lacks stones, yet both the generated left and right views erroneously include them. Similarly, fine details such as text often become distorted. Quantitative reconstruction loss measurements (Tab. ~\ref{tab:reconstruction}) confirm these issues, showing significantly higher errors compared to our method. Moreover, using StereoDiffusion-generated stereo pairs for training stereo matching networks led to poor performance and convergence difficulties.

\begin{table*}
    \centering
    \begin{tabular}{l|cc|cc|cc|cc}
        \hline
        \multirow{2}{*}{Method} & \multicolumn{2}{c|}{Cloudy} & \multicolumn{2}{c|}{Foggy} & \multicolumn{2}{c|}{Rainy} & \multicolumn{2}{c}{Sunny} \\
         & F & H & F & H & F & H & F & H \\ \hline
        NS-RAFT-Stereo & 8.81 & 2.95 & 18.18 & 3.41 & \textbf{29.19} & \textbf{8.47} & 7.42 & \textbf{2.88} \\
        Zero-RAFT-Stereo & \textbf{6.44} & \textbf{2.69} & \textbf{8.66} & \textbf{1.70} & 30.10 & 11.71 & \textbf{6.46} & 3.15 \\ \hline
    \end{tabular}
    \vspace{-5px}
    \caption{Zero-shot generalization performance on DrivingStereo under different weather. We utilize \textgreater 3px All in comparisons.}
    \label{tab:driving}
    \vspace{-5px}
\end{table*}

\begin{figure*}[t]
    \centering
    \includegraphics[width=1.0\textwidth]{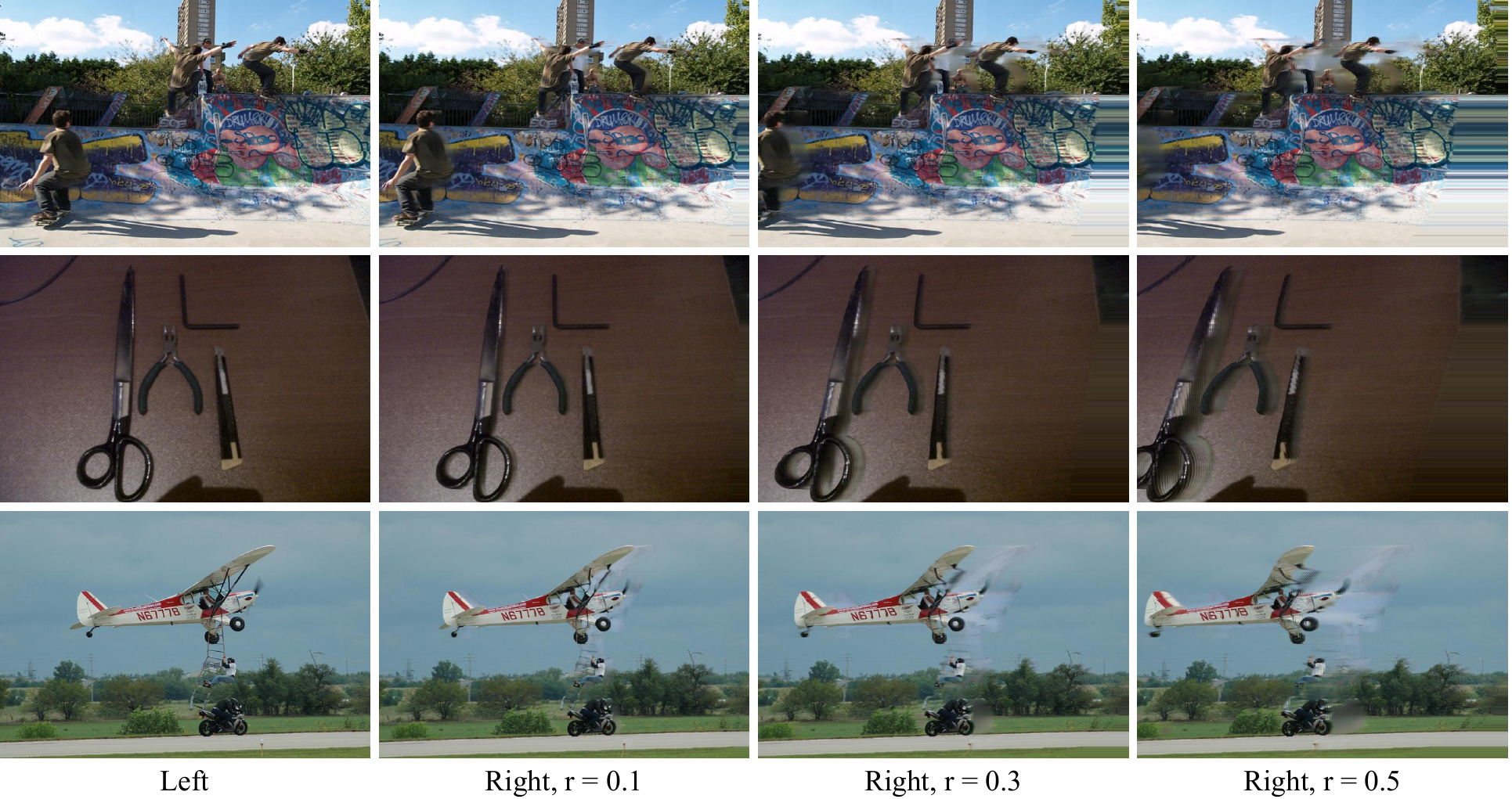}
    \caption{Visualization of AdaMPI~\cite{han2022single}.}
    \label{fig:mpi}
    \vspace{-10px}
\end{figure*}

AdaMPI~\cite{han2022single} generates multiplane images~\cite{zhou2018stereo} (MPI) from a single input image from a single input image for novel view synthesis. However, as shown in Fig. ~\ref{fig:mpi}, varying the camera motion ratios often introduces artifacts, particularly in occluded regions, where ghosting and trailing effects are prevalent. This suggests that the MPI approach struggles to reconstruct the scene's semantic structure accurately. As a result, MPI-based stereo generation is less suitable for training stereo matching models, as these artifacts compromise the quality and consistency needed for effective learning.

In summary, while StereoDiffusion~\cite{wang2024stereodiffusion} and AdaMPI~\cite{han2022single} introduce innovative approaches for synthesizing stereo images from single inputs, both have significant limitations. StereoDiffusion suffers from high computational costs and content distortions, while AdaMPI struggles with semantic inconsistencies in occluded regions. These challenges highlight the need for more robust and accurate synthesis methods for stereo matching applications.

\section{Additional Comparisons with NeRF-Stereo}
\label{sec:benchmark}

In this section, we present additional comparisons with NeRF-Stereo~\cite{tosi2023nerf}, detailed Midd-T benchmark results, visualizations on KITTI and ETH3D, and zero-shot generalization performance on DrivingStereo~\cite{yang2019drivingstereo}.

For Midd-T, we report the performance of each sample in Tab. ~\ref{tab:detail_middt}. Compared to NS-RAFT-Stereo~\cite{tosi2023nerf}, our Zero-RAFT-Stereo achieves improvements in nearly all cases. Notably, for samples where NS-RAFT-Stereo performs poorly, our method improves accuracy by almost 50\%.

\begin{table*} \footnotesize
    \centering
    \begin{tabular}{l|ccccccccccccccc}
        \hline
        Method & Adi. & ArtL & Jad. & Mot. & Mot.E & Pia. & Pia.L & Pip. & Plr. & Plt. & Plt.P & Rec. & She. & Ted. & Vin. \\ \hline
        NS-RAFT-Stereo & 1.51 & \textbf{4.14} & 24.90 & 3.62 & 4.04 & 9.04 & 25.81 & 5.89 & 14.08 & \textbf{6.13} & 5.54 & 4.94 & 39.59 & 4.96 & 26.35 \\
        Zero-RAFT-Stereo & \textbf{1.39} & 4.91 & \textbf{14.27} & \textbf{3.26} & \textbf{3.68} & \textbf{5.69} & \textbf{13.73} & \textbf{5.22} & \textbf{9.53} & 7.21 & \textbf{5.50} & \textbf{4.20} & \textbf{23.97} & \textbf{4.77} & \textbf{18.01} \\ \hline
    \end{tabular}
    \vspace{-5px}
    \caption{Details of Midd-T. We utilize \textgreater 2px Noc regions in Midd-T (F)}
    \label{tab:detail_middt}
    \vspace{-5px}
\end{table*}

\begin{figure*}[t]
    \centering
    \includegraphics[width=1.0\textwidth]{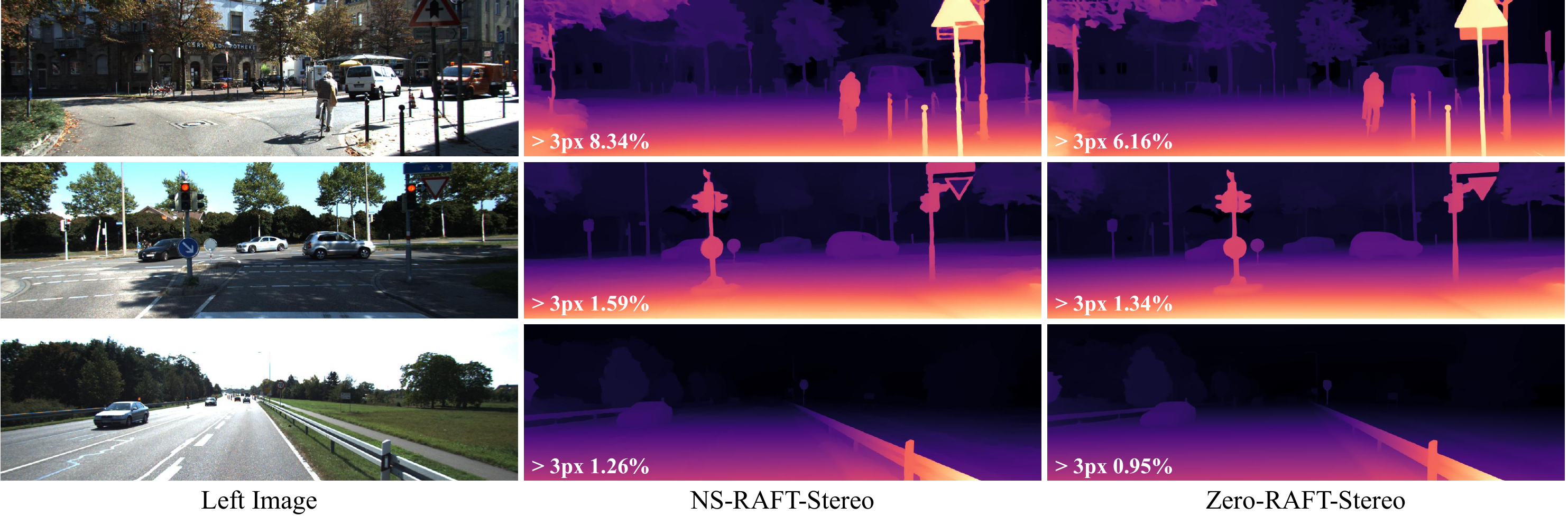}
    \caption{Visualization of KITTI.}
    \label{fig:kitti}
    \vspace{-10px}
\end{figure*}

\begin{figure*}[t]
    \centering
    \includegraphics[width=1.0\textwidth]{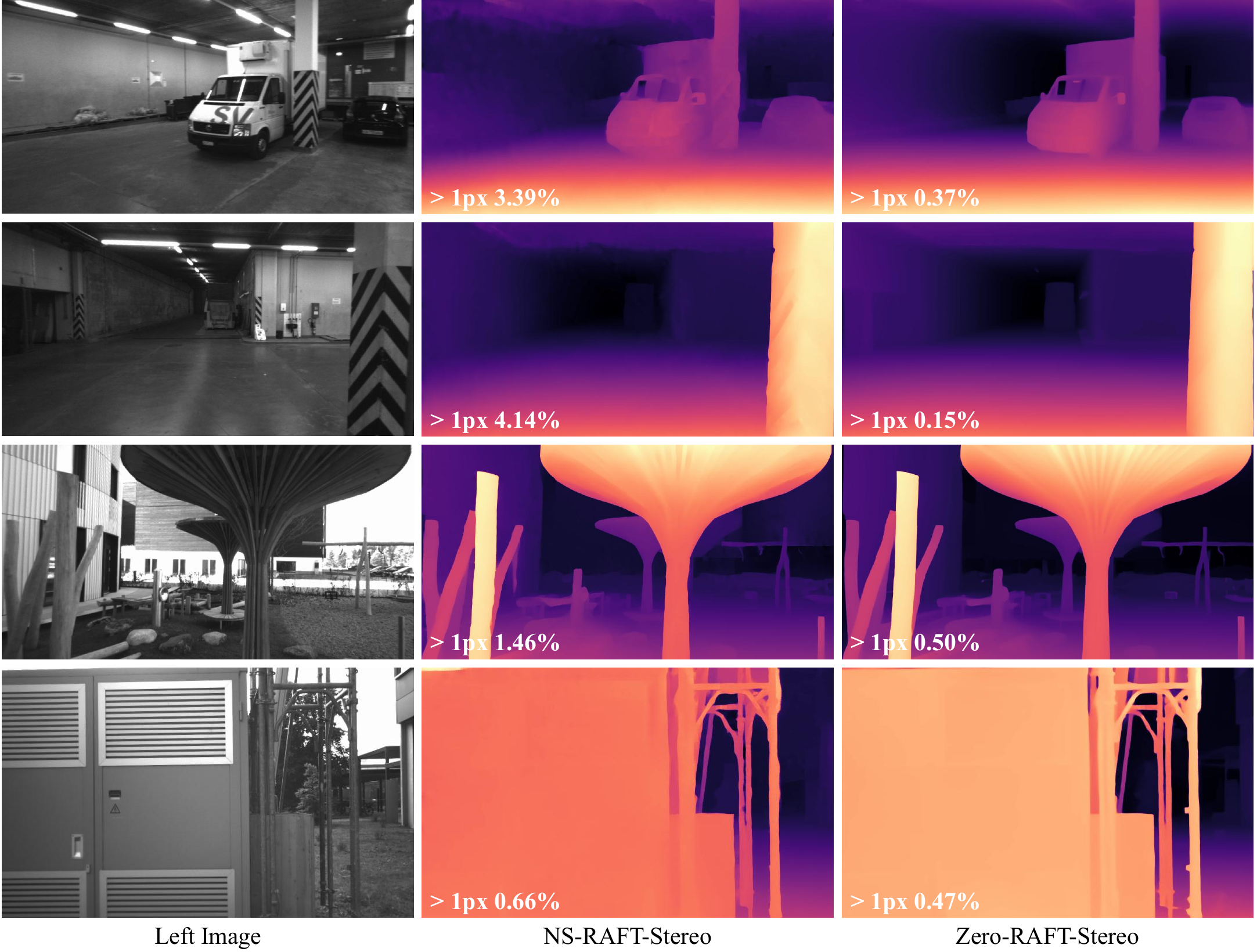}
    \caption{Visualization of ETH3D.}
    \label{fig:eth3d}
    \vspace{-10px}
\end{figure*}

For KITTI and ETH3D, we provide visual comparisons between NS-RAFT-Stereo and Zero-RAFT-Stereo. As shown in Fig. ~\ref{fig:kitti}, Fig. ~\ref{fig:eth3d}, Zero-RAFT-Stereo generates smoother and more accurate disparity maps with fewer artifacts and reduced noise. Notably, in the second row of Fig. ~\ref{fig:eth3d},  our model effectively removes the large disparity artifacts present in NS-RAFT-Stereo, particularly in the central dark region, demonstrating its superior handling of challenging textures and illumination variations.

\begin{figure*}[t]
    \centering
    \includegraphics[width=1.0\textwidth]{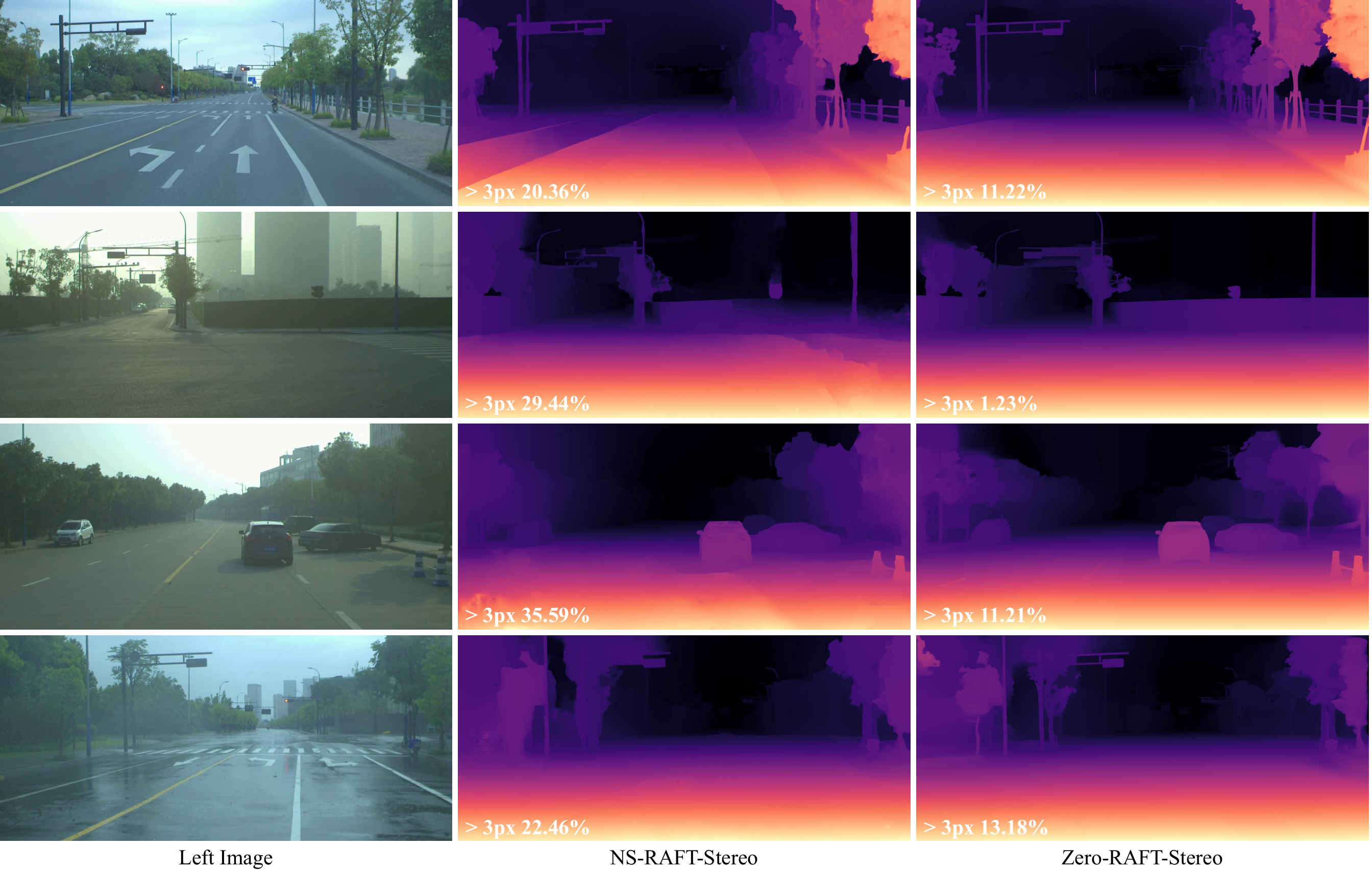}
    \caption{Visualization of DrivingStereo.}
    \label{fig:driving}
    \vspace{-10px}
\end{figure*}

Additionally, we evaluate both models on the DrivingStereo dataset under different weather conditions. As shown in Tab. ~\ref{tab:driving}, our Zero-RAFT-Stereo outperforms NS-RAFT-Stereo across all weather conditions except rainy weather, where both models exhibit poor performance, indicating a need for further optimization in such scenarios. Notably, Zero-RAFT-Stereo demonstrates significant improvements under foggy conditions, reducing errors from 18.18\% to 8.66\% at full resolution and from 3.41\% to 1.70\% at half resolution. Since foggy scenes typically have low contrast and poor visibility, these results suggest that Zero-RAFT-Stereo is more robust in such challenging conditions. As illustrated in Fig. ~\ref{fig:driving}, under extreme weather conditions, NS-RAFT-Stereo struggles to predict large textureless regions, while Zero-RAFT-Stereo successfully reconstructs complete ground surfaces and walls. Moreover, Zero-RAFT-Stereo exhibits superior segmentation of thin, tree-like objects and blurry background regions, highlighting its ability to maintain fine details even in adverse conditions.

\end{document}